\documentclass[letterpaper, 10 pt, conference]{ieeeconf} 
\IEEEoverridecommandlockouts
\usepackage[utf8]{inputenc}
\usepackage{amsfonts}
\usepackage{float}
\usepackage[T1]{fontenc}
\usepackage[noadjust]{cite}

\usepackage [autostyle, english = american]{csquotes}
\MakeOuterQuote{"}
\usepackage{textcomp}
\usepackage{blindtext}
\usepackage{graphicx}
\usepackage{lettrine}
\usepackage{lipsum}
\usepackage{multicol}
\usepackage{multirow}
\usepackage{textcomp}
\usepackage{layouts}
\usepackage[font=small,skip=0pt]{caption}
\usepackage{amsmath}
\usepackage{color}
\usepackage{mathtools}
\usepackage{booktabs}

\usepackage{array}
\newcolumntype{L}[1]{>{\raggedright\let\newline\\\arraybackslash\hspace{0pt}}m{#1}}
\newcolumntype{C}[1]{>{\centering\let\newline\\\arraybackslash\hspace{0pt}}m{#1}}
\newcolumntype{R}[1]{>{\raggedleft\let\newline\\\arraybackslash\hspace{0pt}}m{#1}}
\usepackage[linesnumbered]{algorithm2e}
\usepackage{bbm}
\usepackage{duckuments}
\usepackage{hyperref}
\hypersetup{colorlinks,linkcolor={black},citecolor={blue},urlcolor={blue}}
\usepackage[noadjust]{cite}
\usepackage{subcaption}
\usepackage{array}
\newcolumntype{P}[1]{>{\centering\arraybackslash}p{#1}}
\usepackage{balance}
\usepackage{epstopdf}

\title{\LARGE \bf
Multi-Model Predictive Attitude Control of Quadrotors}
\author{Mohammadreza Izadi, Zeinab Shayan, and Reza Faieghi
\thanks{*This work was partially supported by the Natural Sciences and Engineering Research Council of Canada (NSERC).}
\thanks{The authors are with the Autonomous Vehicles Laboratory, Department of Aerospace Engineering, Toronto Metropolitan University, Toronto, Canada.
        {\tt\small \{mizadi, zshayan, reza.faieghi\}@torontomu.ca}}
}

\begin{document}
\maketitle
\thispagestyle{empty}
\pagestyle{empty}
\bstctlcite{IEEEexample:BSTcontrol}

\begin{abstract}
This paper introduces a new multi-model predictive control (MMPC) method for quadrotor attitude control with performance nearly on par with nonlinear model predictive control (NMPC) and computational efficiency similar to linear model predictive control (LMPC). Conventional NMPC, while effective, is computationally intensive, especially for attitude control that needs a high refresh rate. Conversely, LMPC offers computational advantages but suffers from poor performance and local stability. Our approach relies on multiple linear models of attitude dynamics, each accompanied by a linear model predictive controller, dynamically switching between them given flight conditions. We leverage gap metric analysis to minimize the number of models required to accurately predict the vehicle behavior in various conditions and incorporate a soft switching mechanism to ensure system stability during controller transitions. Our results show that with just 15 models, the vehicle attitude can be accurately controlled across various set points. Comparative evaluations with existing controllers such as incremental nonlinear dynamic inversion, sliding mode control, LMPC, and NMPC reveal that our approach closely matches the effectiveness of NMPC, outperforming other methods, with a running time comparable to LMPC.
\end{abstract}

\section{INTRODUCTION}\label{se:intro}
Improving the flight controller of quadrotors allows them to navigate more effectively in complex environments.
This extends the versatility of quadrotors, and paves the way for more sophisticated and autonomous operations, extending their applications. 
However, flight control of quadrotors faces several challenges.

One challenge is the underactuation of quadrotors which leads to complexity in the control system architecture \cite{mahony2012multirotor}.
A typical flight controller involves a cascaded structure, comprising a position controller in the outer loop and an attitude controller in the inner loop \cite{yazdanshenas2024quaternion}. 
To ensure stability, the attitude controller must operate at significantly higher frequencies than the position controller.
For instance, in the PX4 Autopilot system, the attitude controller runs at 250 Hz, compared to the 50 Hz of the position controller \cite{PX4ControllerDiagrams2024}. 

Another challenge is the limited computing power of quadrotors, especially in nano (<250 $[g]$) and micro (250 $[g]$ -- $2 [kg]$) classes.
This severely restricts the sophistication of the flight controller, and in turn, the maximum maneuverability of the vehicle, highlighting the role of computational efficiency in flight control algorithms.
As the push towards more autonomous operations demands a higher computational load on onboard processors, achieving high-performance and fast control algorithms becomes an important research focus.

One control method that has proven effective in the advanced control of quadrotors is model predictive control (MPC), offering optimization-based control while explicitly handling system constraints.
Several studies have applied linear MPC (LMPC) for quadrotor control \cite{bangura2014real, alexis2012model, shayan2024nonlinear}.
The LMPC is simple and computationally efficient, but since it relies on a linearized model of the quadrotor, it has limited performance.
Nonlinear MPC (NMPC) alleviates this issue but at the cost of significantly higher computational loads.
As a result, the application of NMPC for quadrotors has been primarily limited to only position control \cite{sun2022comparative, nan2022nonlinear}, or only attitude control \cite{kamel2015fast}.
In \cite{wang2021efficient}, NMPC is applied for both position and attitude control, but the flight envelope is limited to small pitch and roll and only yaw is controlled.

One variant of MPC that compromises the computational efficiency of LMPC and the performance of NMPC is Multi-Model Predictive Control (MMPC) \cite{di2004multi}. This method incorporates a set of linear models, each for specific operating conditions or system uncertainty scenarios. 
Compared to LMPC, MMPC predicts system behavior more accurately by dynamically selecting the most appropriate model for the current conditions.
Finding the control input boils down to solving a constrained linear optimization problem which is significantly faster than solving nonlinear problems in NMPC.
Therefore, MMPC has the potential to offer higher performance than LMPC, while still being computationally more efficient than NMPC.

In light of the above discussion, we focus on MMPC design for quadrotors.
We target the attitude control loop as it will benefit the most from a computationally efficient controller given its higher framerate than the position control loop.
However, as the position controller depends on attitude control, our performance improvements in the attitude control will boost the position control accuracy as well.
MMPC design for nonlinear systems is an active area of research.
For recent work on MMPC, we cite \cite{liang2020holistic,pipino2021adaptive, sun2023energy, ma2024multi, liumulti}.
The existing studies in the context of quadrotors include \cite{alexis2011switching, li2022enhanced}.

One formidable challenge in the design of MMPC lies in the creation of a linear model bank.
On one hand, it is beneficial to increase the number of models to account for more operating and uncertainty conditions.
On the other hand, each model requires a new controller calibration, testing, and validation, making the control development process effortful.
Inspired by \cite{du2014gap}, we employ gap metric analysis to find the minimum number of linear models to be included in the model bank. This keeps the overall structure of MMPC simple while covering various dynamical characteristics over the total flight envelope.

Another challenge in MMPC design is poor performance during the transition from one controller to another. 
An abrupt change in the controller can lead to an abrupt change of control input, actuator failure, unexpected large peaks in the system states, and even instability \cite{ma2024multi}.
This can be handled using soft-switching algorithms that create a smooth transition from one controller to another \cite{wang2005analysis, saki2018optimal, wang2006softly}. We adopt the method presented in \cite{wang2005analysis} to create a soft-switching mechanism, ensuring system stability and high performance.

It is worth noting that our approach lends well to various common position controllers. In this paper, we use sliding mode control (SMC) due to its simplicity. Other controllers can potentially be considered for position control, depending on specific application requirements and system characteristics. This flexibility allows for exploring alternative nonlinear control approaches to enhance the performance and robustness of quadrotor trajectory control while benefiting from an MPC-based attitude control.

Overall, the major contribution of this work is to develop MMPC for quadrotor attitude control, striking a balance between the performance of NMPC, and the computational efficiency of LMPC. The key strengths of our work compared to the existing studies \cite{alexis2011switching, li2022enhanced} include:
\begin{itemize}
    \item Using gap metric to minimize the number of models required in the model bank without affecting the number of operating points and/or uncertainty scenarios considered.
    \item Developing soft-switching mechanisms to ensure the stable transition between controllers. 
\end{itemize}

\section{QUADROTOR DYNAMICS}
\begin{figure}[t]
    \centering
    \includegraphics[width = 1\linewidth]{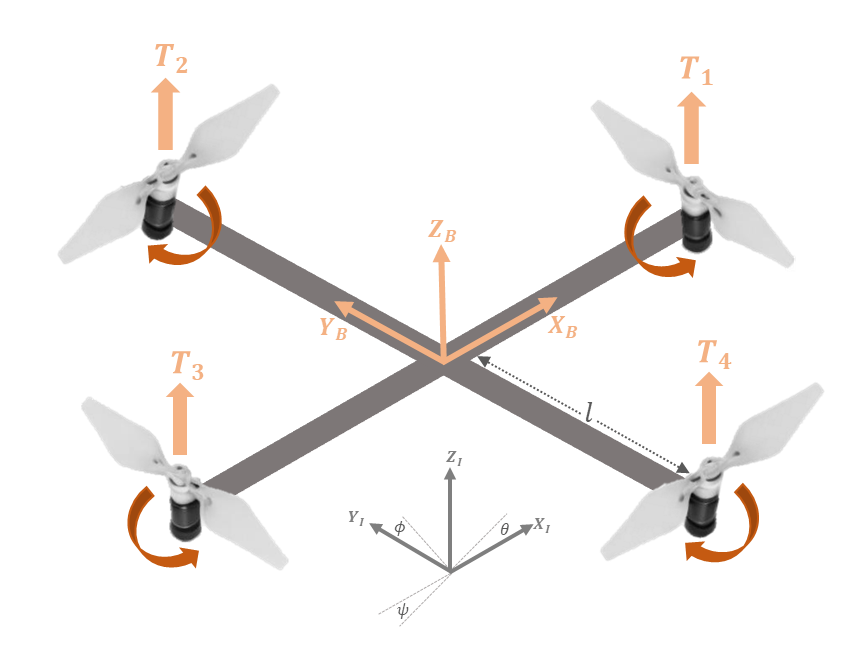}
    \caption{Quadrotor model and coordinate frames}
    \label{fig:modeling}
\end{figure}

This section presents the equations of flight for quadrotors to establish the notation for our control developments.
Let us begin by setting $\mathcal{I}=\left\{{\bf{x}}_\mathcal{I},{\bf{y}}_\mathcal{I},{\bf{z}}_\mathcal{I} \right\}$ as an Earth-fixed inertial coordinates frame, and $\mathcal{B}=\left\{{\bf{x}}_\mathcal{B},{\bf{y}}_\mathcal{B},{\bf{z}}_\mathcal{B} \right\}$ as the body-fixed coordinates frame whose origin coincides with the center of mass of the quadrotor (Fig. \ref{fig:modeling}). 
We assume that the quadrotor body is rigid and symmetric, with arms aligned to ${\bf{x}}_\mathcal{B}$ and ${\bf{y}}_\mathcal{B}$.
The length of each arm is $l$, the mass of the vehicle is $m$, and the inertia matrix is ${\bf{J}}$ which is diagonal ${\bf{J}} = {\rm{diag}}\left(J_x,J_y,J_z\right)$ due to the symmetry of the vehicle.

We denote the position and velocity of the vehicle in $\mathcal{I}$ by ${\boldsymbol{\xi}}=\left[x,y,z\right]^T$ and $\mathbf{v}=\left[u,v,w\right]^T$, respectively.
For the vehicle attitude, we use ${\boldsymbol{\eta}}=\left[\phi, \theta,\psi\right]^T$, where $ - \pi  < \phi  \le \pi $, $ - \frac{\pi }{2} \le \theta  \le \frac{\pi }{2}$, and $ - \pi  < \psi  \le \pi $ are the Euler angles representing pitch, roll, and yaw in the yaw-pitch-roll sequence.
With the above Euler angle configuration, the rotation matrix from $\mathcal{B}$ to $\mathcal{I}$ takes the following form
\begin{equation}
\label{eq:dy1}
{\bf{R}} = \left[ {\begin{array}{*{20}{c}}
{c\theta c\psi }&{s\phi s\theta c\psi  - c\phi s\psi }&{c\phi s\theta c\psi  + s\phi s\psi }\\
{c\theta s\psi }&{s\phi s\theta s\psi  + c\phi c\psi }&{c\phi s\theta s\psi  - s\phi c\psi }\\
{ - s\theta }&{s\phi c\theta }&{c\phi c\theta }
\end{array}} \right],
\end{equation}
where $c$ and $s$ stand for cosine and sine functions.
Also, if ${\boldsymbol{\omega}}=\left[p,q,r\right]^T$ represents the angular velocity vector, then according to the Euler kinematical equation, we have $\dot {\boldsymbol{\eta}} = {\bf{H}}\left({\boldsymbol{\eta}}\right) {\boldsymbol{\omega}}$, where 
\begin{equation}\label{eq:dy2}
{\bf{H}}\left( {\boldsymbol{\eta }} \right) = \left[ {\begin{array}{*{20}{c}}
1&{\sin \phi \tan \theta }&{\cos \phi \tan \theta }\\
0&{\cos \phi }&{ - \sin \phi }\\
0&{{{\sin \phi } \mathord{\left/
 {\vphantom {{\sin \phi } {\cos \theta }}} \right.
 \kern-\nulldelimiterspace} {\cos \theta }}}&{{{\cos \phi } \mathord{\left/
 {\vphantom {{\cos \phi } {\cos \theta }}} \right.
 \kern-\nulldelimiterspace} {\cos \theta }}}
\end{array}} \right].
\end{equation}

Each of the vehicle's rotors produces a thrust $T_i,\;i=\left\{1,2,3,4\right\}$ in the direction of $\bf{z}_\mathcal{B}$. 
$T_i$s are usually approximated by $k_T \Omega_i^2$ where $\Omega_i$ is the angular velocity of $i$-th rotor, and $k_T$ is a coefficient.
The rotor angular velocities on the $\bf{x}_\mathcal{B}$ and $\bf{y}_\mathcal{B}$ axes have opposite signs ($\Omega_{1,3} > 0$, $\Omega_{2,4} < 0$) to counterbalance the reaction torque induced by the rotors and to control $\psi$.
Let ${\bf{f}}$ and ${\boldsymbol{\tau}}$ be the aerodynamic force and torque vectors produced by $T_i$s.
Then, we can express them in $\mathcal{B} $ as follows
\begin{equation}\label{eq:dy3}
{\bf{f}} = \left[ {\begin{array}{*{20}{c}}
0\\
0\\
T
\end{array}} \right],\;{\rm{and}}\;{\boldsymbol{\tau}} = \left[ {\begin{array}{*{20}{c}}
{lk_T\left( {\Omega _2^2 - \Omega _4^2} \right)}\\
{lk_T\left( {\Omega _3^2 - \Omega _1^2} \right)}\\
{k_Q\left( {-\Omega _1^2 + \Omega _2^2 - \Omega _3^2 + \Omega _4^2} \right)}
\end{array}} \right],
\end{equation}
where $l$ is the distance from the center of mass to the rotor, $T={k_T}\sum\nolimits_{i = 1}^4 {\Omega _i^2}$ is the total thrust, $k_T$ is thrust coefficient and $k_Q$ is torque coefficient. 

Applying Newton's law of motion, the translational dynamics of the vehicle take the following form
\begin{equation}\label{eq:dy4}
    {\ddot {\boldsymbol{\xi}}  =  - {{\bf{g}}} + \frac{1}{m}\mathbf{Rf}},
\end{equation}
where ${{\bf{g}}} = \left[0\;0\;g\right]^T$ is the gravity vector with $g$ set to $9.81\;[m/s^2]$.
Using Euler's rotation theorem, the rotational dynamics of the vehicle take the following form
\begin{equation}\label{eq:dy5}
    \dot{\boldsymbol{\omega}} = {\bf{J}}^{ - 1}\left( -{\boldsymbol{\omega}} \times {\bf{J}}{\boldsymbol{\omega}} + {\boldsymbol{\tau}} \right),\;{\rm{and}}\;\dot {\boldsymbol{\eta}}  = {\bf{H}}\left( {{\boldsymbol{\eta}} } \right){\boldsymbol{\omega}}.
\end{equation}
To develop MMPC, we will linearize \eqref{eq:dy5} around various operating points.



\section{CONTROLLER DESIGN}
\begin{figure*}[t]
    \centering
    \includegraphics[trim={0cm, 5.5cm, 0.2cm, 2.5cm}, clip, width = 0.8\linewidth]{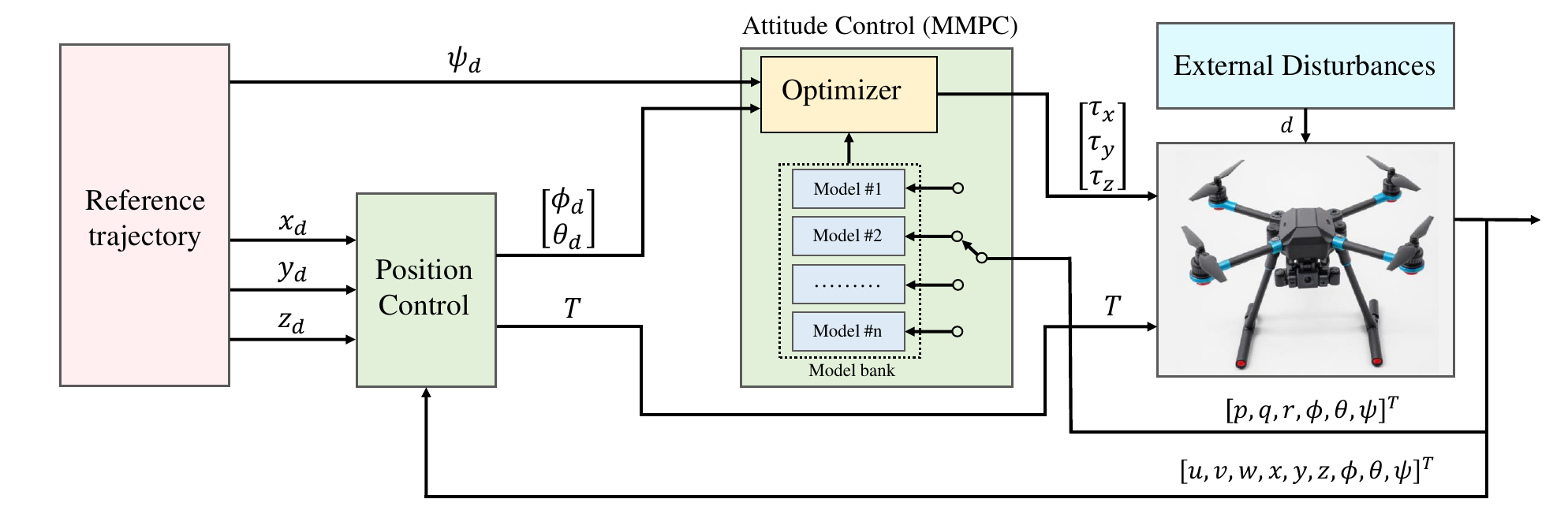}
    \caption{Control system block diagram}
    \label{fig:block_diagram}
\end{figure*}

Figure \ref{fig:block_diagram} illustrates our flight control architecture.
We consider a cascaded structure with the proposed MMPC as the attitude controller.

As is the case with most quadrotor flight control architectures, the desired roll and pitch of the attitude controller are determined by the position controller, while the desired yaw is given by the user. 
Note that the position controller itself can be another MMPC formulated based on the \eqref{eq:dy4}, or other control algorithms.
In this work, we will adopt an existing sliding mode control (SMC) law \cite{IZADI2024105854} for position control, and focus this section on attitude control with MMPC.

\subsection{Model Bank}
The initial step in MMPC design is to construct the model bank.
To this end, we choose $M$ operating points, evenly distributed in the system state space, and linearize the attitude dynamics \eqref{eq:dy5} around each operating point.
The models will take the following form
\begin{equation}\label{eq:mpc1}
\boldsymbol{\chi}_m(k + 1) = \mathbf{A}_m\boldsymbol{\chi}_m(k) + \mathbf{B}_m \mathbf{u}_m(k),
\end{equation}
where $1 \le m \le M$ is the model index, ${\boldsymbol{\chi}_m} = [\Delta \phi,\Delta \theta,\Delta \psi, \Delta p,\Delta q,\Delta r]^T$, $\mathbf{u}_m = {[\Delta {\tau_x},\Delta {\tau_y},\Delta {\tau_z}]^T}$, 
$\mathbf{A}_m \in \mathbb{R}^{6\times6}$ and $\mathbf{B}_m\in\mathbb{R}^{6\times3}$.

Note that, in building the model bank, we apply the state constraints $- \pi  < \phi  \le \pi $, $ - \frac{\pi }{2} \le \theta  \le \frac{\pi }{2}$ per Euler angle definitions.
Also, during linearization, the column of the Jacobian matrix that corresponds to $\psi$ remains zero in all operating points; therefore, the linear models at any given point remain independent of $\psi$.

As shown in Fig. \ref{fig:block_diagram}, at each iteration, the MMPC compares the current values of $\phi$ and $\theta$ with the ensemble of operating points, identifies the nearest operating point, and selects the corresponding model for MPC calculations.

\subsection{Model Bank Reduction Using Gap Metric Analysis}
As mentioned in Section \ref{se:intro}, having a larger number of models can improve performance by better matching the system's behavior under different conditions; however, this leads to tedious control calibrations and may strain the system's memory resources.
One of the key features of our design is to adopt gap metric analysis to minimize the number of models needed in the model bank.

The gap metric is a quantitative measure of dissimilarity between two linear models.
If the differences among several models in the model bank fall below a specified threshold, then a single model is sufficient to address all the scenarios for which those models were originally intended.
Let us consider two systems with transfer functions $G_1\left(s\right)=N_1\left(s\right)D_1^{-1}\left(s\right)$ and $G_2\left(s\right)=N_2\left(s\right)D_2^{-1}\left(s\right)$, where $N_i$ and $M_i$ are normalized coprime factors.
According to \cite{el1985gap}, the directed distance from $G_1$ to $G_2$, denoted by $\vec{\delta}\left(G_1,G_2\right)$, is the smallest difference across all stable compensators $Q\left(s\right)$ that could align $G_2$ with $G_1$, given by
\begin{equation}\label{eq:delta2}
\vec{\delta}\left(G_1,G_2\right)={\min_{Q \in H_\infty}}{\left\| {\left[ {\begin{array}{*{20}{c}}
{{D_1}}\\
{{N_{1}}}
\end{array}} \right] - \left[ {\begin{array}{*{20}{c}}
{{D_2}}\\
{{N_2}}
\end{array}} \right]Q} \right\|_\infty},
\end{equation}
where $H_\infty$ is a Hardy space of transfer functions that are bounded and analytic in the right half of the complex plane ensuring stable compensators, and $\| \cdot \|_\infty$ is the $H_\infty$ norm defined as the maximum singular value of transfer function over all frequencies.
In general, $\vec{\delta}\left(G_1,G_2\right) \neq \vec{\delta}\left(G_2,G_1\right)$. 
As such, the gap metric between $G_1$ and $G_2$ is defined as
\begin{equation}\label{eq:delta1}
    \delta\left(G_1,G_2\right) = \max \left\{ \vec{\delta}\left(G_1, G_2\right) , \vec{\delta}\left(G_2,G_1\right) \right\}.
\end{equation}

The gap metric satisfies $ 0\le\delta\left(G_1, G_2 \right)\le1$.
Values close to zero indicate that the frequency responses of $G_1$ and $G_2$ are similar across all frequencies. Thus, a controller designed based on one system's characteristics is likely to achieve similar performance when applied to the other system.
We define a threshold $\delta_{th}$, and for any two systems whose gap metric is smaller than $\delta_{th}$, we keep only one of the systems in the model bank.
Therefore, from the initially $M$ linear models, we will have $M'$ number of models left in the model bank.



\subsection{MPC Formulation}
We use the standard LMPC formulation to design a control law for each model remaining in the model bank.
Let $N$ be the prediction horizon, $\mathbf{P}_m, \mathbf{Q}_m \in \mathbb{R}^{6 \times 6}$ positive semi-definite, and $\mathbf{R}_m \in \mathbb{R}^{3 \times 3}$ positive definite. Then, the control law for the $m$-th model is the solution to the following constrained optimization problem
\begin{equation}\label{eq:mpc2}
\begin{aligned}
    \min J_m(k) = & \boldsymbol{\chi}_m^T(k+N|k)\mathbf{P}_m\boldsymbol{\chi}_m(k+N|k) \\
& + \sum_{i=0}^{N-1} \left[ \boldsymbol{\chi}_m^T(k+i|k) \mathbf{Q}_m \boldsymbol{\chi}_m(k+i|k) \right. \\
& \left. + \mathbf{u}_m^T(k+i|k) \mathbf{R}_m \mathbf{u}_m(k+i|k) \right],
\end{aligned}
\end{equation}
subject to
\begin{equation}\label{eq:mpc3}
\begin{aligned}
\boldsymbol{\chi}_m\left(k|k\right)&=\boldsymbol{\chi}_m\left(k\right)\\
\boldsymbol{\chi}_m\left(k+i+1|k\right)&=\mathbf{A}_m\boldsymbol{\chi}_m\left(k+i|k\right)+\mathbf{B}_m \mathbf{u}_m\left(k+i|k\right),\\
\boldsymbol{\chi}_m\left(k + i|k\right)& \in \mathcal{X},\forall i=0,\cdots,N-1,\\
\mathbf{u}_m\left(k + i|k\right)& \in \mathcal{U},\forall i=0,\cdots,N-1,\\
\end{aligned}
\end{equation}
where $\mathcal{X}$ and $\mathcal{U}$ are the sets representing acceptable state values and control inputs.

\subsection{Soft Switching}
In most maneuvers, the quadrotor attitude dynamics will undergo large variations, and this necessitates switching between different models and their corresponding control parameters to ensure appropriate control response. 
To effectively handle these transitions, we adopt the soft switching technique presented in \cite{wang2005analysis}.

To this end, let us assume that the switching from the previous controller $1$ to the new controller $2$ starts at $k=k_s$ and completes in $N$ time steps.
In this period, the control parameters will be a weighted sum of the parameters of the two controllers as follows
\begin{equation}
\begin{aligned}
    &\mathbf{P}_{1\rightarrow2} = \alpha_k\left(i\right)\mathbf{P}_1 +\beta_k\left(i\right)\mathbf{P}_2,\\
    &\mathbf{Q}_{1\rightarrow2} = \alpha_k\left(i\right)\mathbf{Q}_1 + \beta_k\left(i\right)\mathbf{Q}_2,\\
    &\mathbf{R}_{1\rightarrow2} = \alpha_k\left(i\right)\mathbf{R}_1 + \beta_k\left(i\right)\mathbf{R}_2,
\end{aligned}
\end{equation}
where $\alpha_k$ and $\beta_k$ are weighting factors, and $0 \le i \le N-1$ is the time sample in the prediction horizon as specified in \eqref{eq:mpc2}.
$\forall\;i\;\in\;\left[0,N-1\right]$, $\alpha_k$ takes the following form
\begin{equation}
    \alpha_k(i) = 
\begin{cases} 
\lambda^{k-k_s+i} & \text{if } k+i < N+k_s, \\
0 & \text{if } k+i \ge N+k_s,
\end{cases}
\end{equation}
where $0 \le \lambda \le 1$ is a design parameter. 
For $i=N$, $\alpha_k\left(N\right) = \alpha_k \left(N-1\right)$.
Furthermore, $\forall\;i\;\in\;\left[0,N\right]$, $\beta_k\left(i\right)$ is described by 
\begin{equation}
\beta_k\left(i\right)=1-\alpha_k\left(i\right).
\end{equation}
According to \cite{wang2005analysis}, the above choice of parameters will ensure system stability during switching.
The key parameter here is $\lambda$ which controls the rate of transition. 
If $\lambda = 0$, the above algorithm will lead to an abrupt switch between the controllers.

\section{RESULTS}
This section presents our simulation results.
We compare the attitude control performance of MMPC with LMPC, NMPC, and two other common attitude control techniques, incremental nonlinear dynamic inversion (INDI) \cite{sun2022comparative} and SMC \cite{IZADI2024105854}.
Given the cascaded flight control structure, the attitude control performance directly impacts position control.
Therefore, we run an additional set of experiments to evaluate the effect of attitude controllers on position tracking while working with identical position controllers.
For the position controller, we use the sliding mode control law developed in our previous work \cite{IZADI2024105854}.

The parameters values for the quadrotor under consideration are $m=650\;[g]$, $\mathbf{J}=\operatorname{diag}\left(0.021,0.023,0.032\right)\;[kg\cdot m^2]$, $l=0.225\;[m]$, $k_T=1.22 \times 10^{-5}$, and $k_Q=689 \times 10^{-7}$.

As for the MMPC, we started with $M=100$ models, linearizing the attitude dynamics around operating points that were evenly distributed in the state space.
Next, we applied our gap metric analysis with $\delta_{th}=0.2$, reducing the model bank to only 15 models. 
For the LMPC formulation of all these models, we set $N=5$, $\mathbf{P}_m = \mathbf{Q}_m = \operatorname{diag}\left(10,10,10,150,150,150\right)$, and $\mathbf{R}_m = \mathbf{I_{3\times3}}$.
Also, we set the soft switching parameter to $\lambda = 0.5$.

\subsection{Scenario 1: Attitude Control Comparison}
In this scenario, the vehicle starts at the origin and adjusts its orientation according to different set points. 
Figure \ref{fig:angular} illustrates the vehicle attitude response using different controllers.
Given our focus on agile flight, we tuned all controllers to achieve their fastest possible settling time.
While all controllers provide fast convergence, their performance varies significantly.
Generally, SMC presents the slowest response, and INDI exhibits overshoots.
The MPC-based controllers deliver smaller errors. NMPC is the quickest, but it has large overshoots in the pitch angle.

For an objective assessment of the controllers, we measure the root mean square (RMS) of errors and the control signals for each controller and tabulate them in Tab. \ref{tab:trajectory_tracking_error2}.
The NMPC is the most accurate, followed closely by the MMPC.
Interestingly, even LMPC performs better than INDI and SMC in minimizing errors, highlighting the advantages of MPC-based strategies over others.

Concerning the control effort, INDI uses the least energy; however, in our experience, it was the most challenging to calibrate, showing high-sensitivity parameter changes.
Among the MPC-based options, MMPC requires the least effort while still maintaining a lower error compared to LMPC, nearly as good as NMPC.

All MPC-based controllers, including MMPC, demonstrated consistent performance.
Given the aim to improve the vehicle's agility, we prioritized $\mathbf{P}_m$ and $\mathbf{Q}_m$ over $\mathbf{R}_m$. However, if reducing control effort is the goal, these parameters can be adjusted accordingly.

\begin{table}[t]
\centering 
  \caption{RMS values for attitude tracking errors and control input}
  \label{tab:trajectory_tracking_error2}
  \small 
  \resizebox{1\linewidth}{!}{
    \begin{tabular}{@{}cccccc@{}}
      \toprule  
      Parameter & MMPC  & LMPC   & NMPC   & INDI   & SMC \\
      \midrule
      $\phi$   & 0.0313 & 0.0517 & 0.0242 & 3.4158 & 5.1953 \\
      $\theta$ & 0.0263 & 0.0451 & 0.0447 & 3.4690 & 4.9202 \\
      $\psi$   & 0.1494 & 0.1948 & 0.0902 & 8.8996 & 11.3058 \\ \midrule

      $\tau_x$ & 0.1517 & 0.0962 & 0.2396 & 0.0262 & 0.1620 \\
      $\tau_y$ & 0.1587 & 0.2165 & 0.2531 & 0.0264 & 0.1743 \\
      $\tau_z$ & 0.0329 & 0.1748 & 0.0546 & 0.0521 & 0.0847 \\ \bottomrule
    \end{tabular}
  }
\end{table}

\begin{figure*}[t]
    \centering
    \includegraphics[trim=3cm 1cm 3cm 0cm,clip, width = \linewidth]{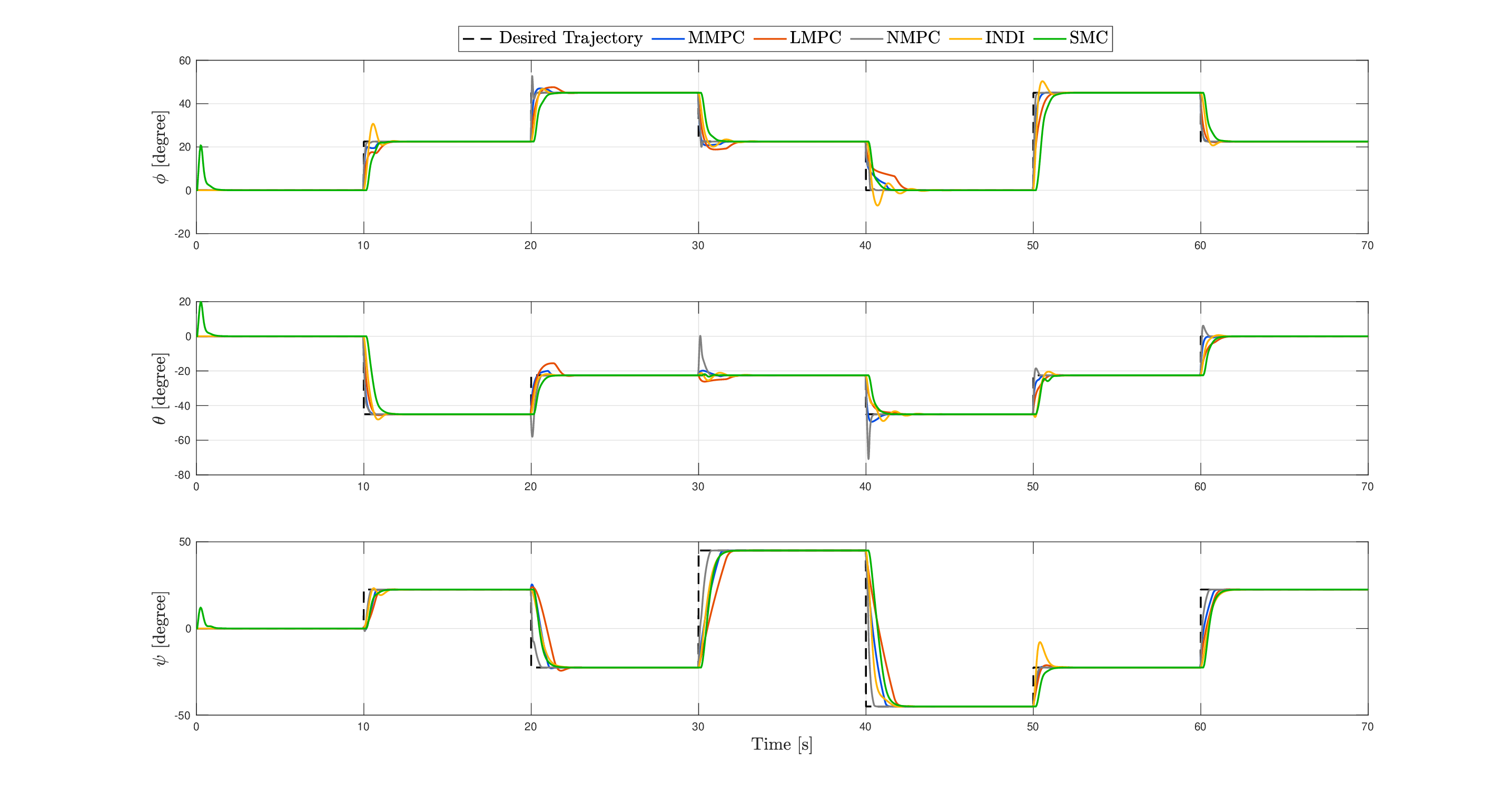}
    \caption{Quadrotor attitude with different controllers}
    \label{fig:angular}
\end{figure*}

\subsection{Scenario 2: Position Control Comparison}
In this scenario, we explore the effects of improved attitude control on position tracking.
The vehicle starts at the initial position $\boldsymbol{\xi}_0 = [0, 0, 0]^T$ [m], and attitude $\boldsymbol{\eta}_0 = \left[0,0,0\right]^T$, tracking a desired trajectory $\boldsymbol{\xi}_d = [5 \sin(0.5t), 5 \cos(0.5t), t]^T$ and desired yaw alternating between $0$, $50^\circ$, and $-50^\circ$.

\begin{figure}[t]
    \centering
    \includegraphics[width = 1\linewidth]{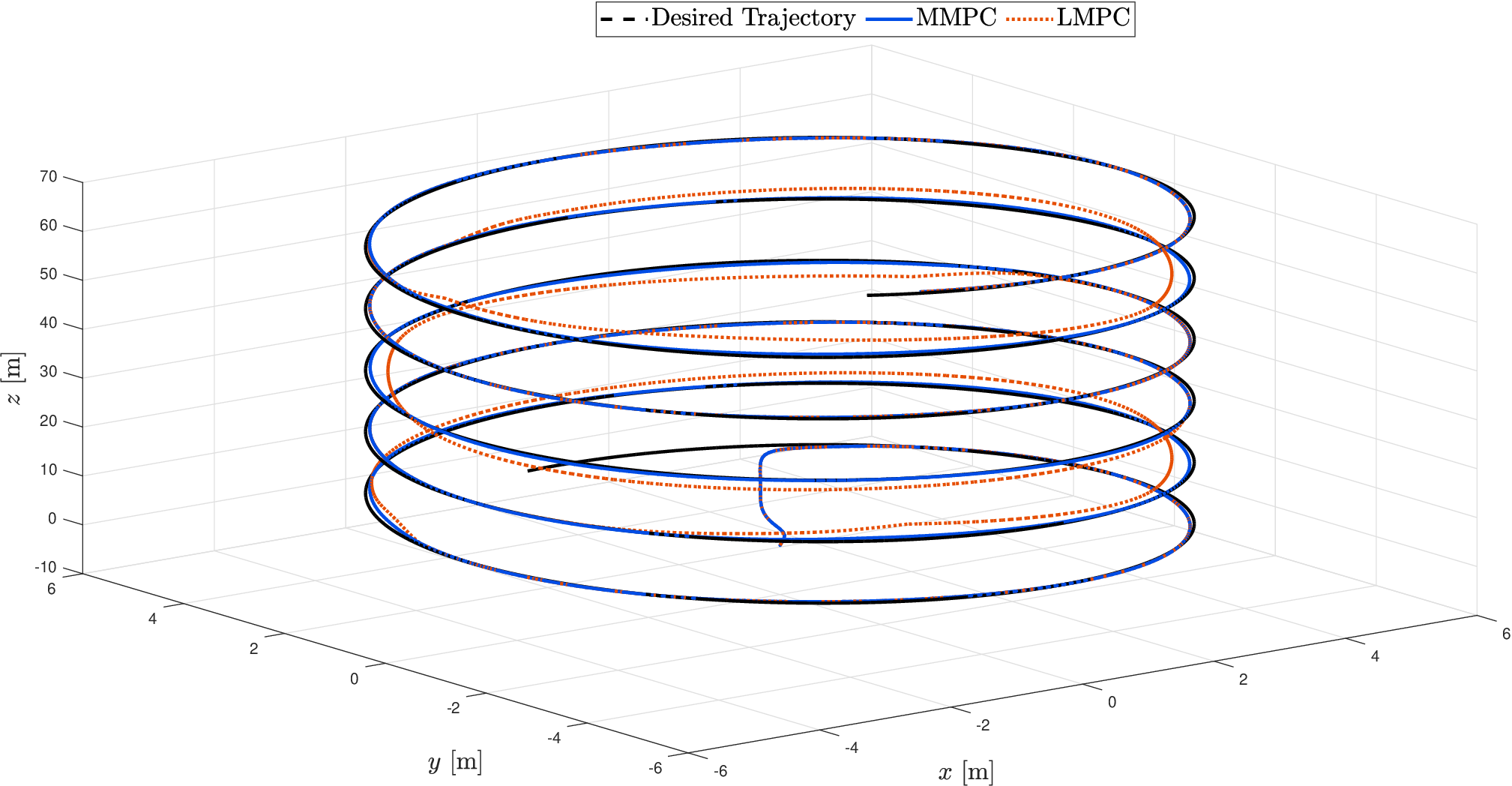}
    \caption{3D plot of the quadrotor trajectory in position control experiments}
    \label{fig:3D}
\end{figure}

\begin{figure}[t]
    \centering
    \includegraphics[width = 1\linewidth]{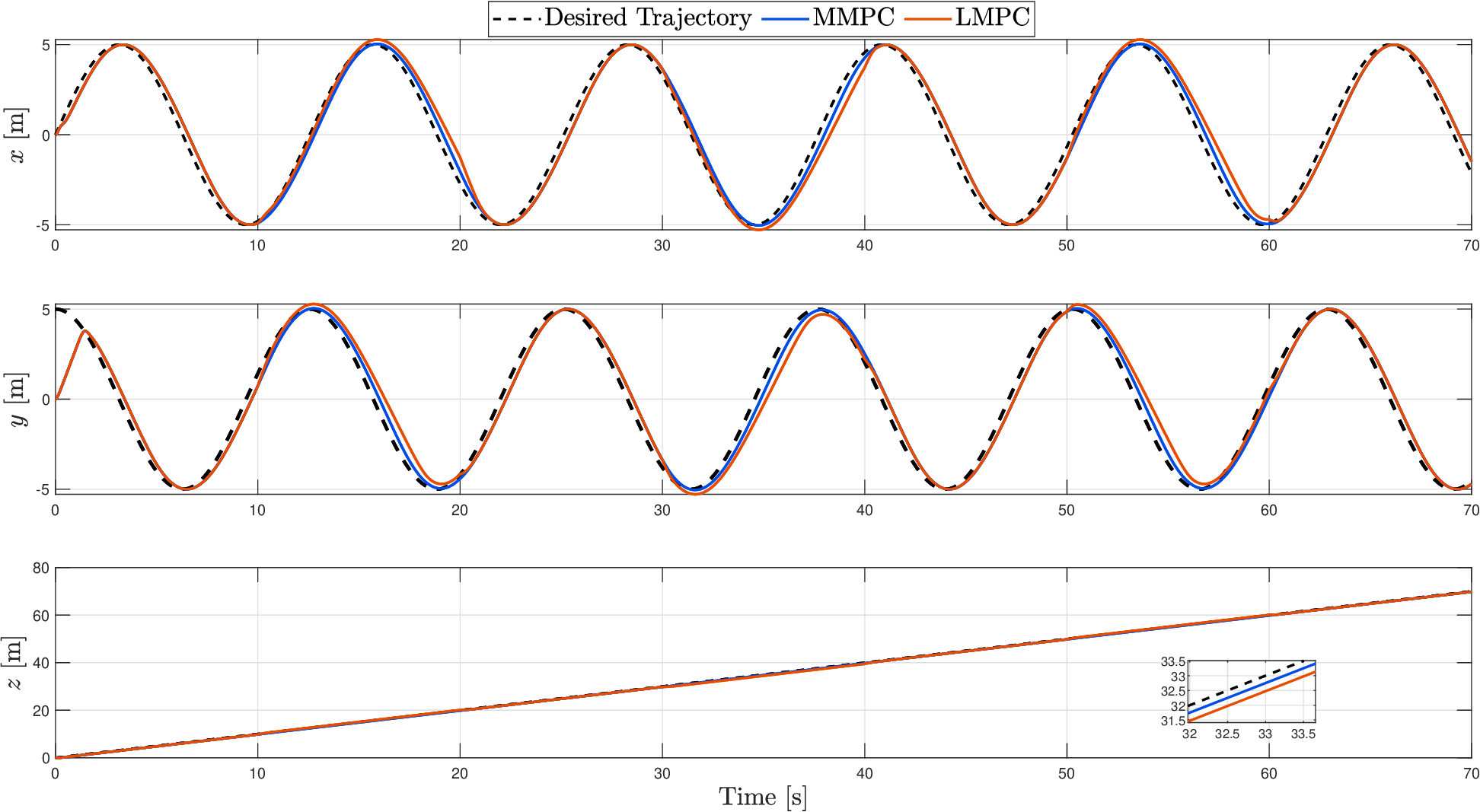}
    \caption{Quadrotor position trajectory in position control experiments}
    \label{fig:position}
\end{figure}

\begin{figure}[t]
    \centering
    \includegraphics[width = 1\linewidth]{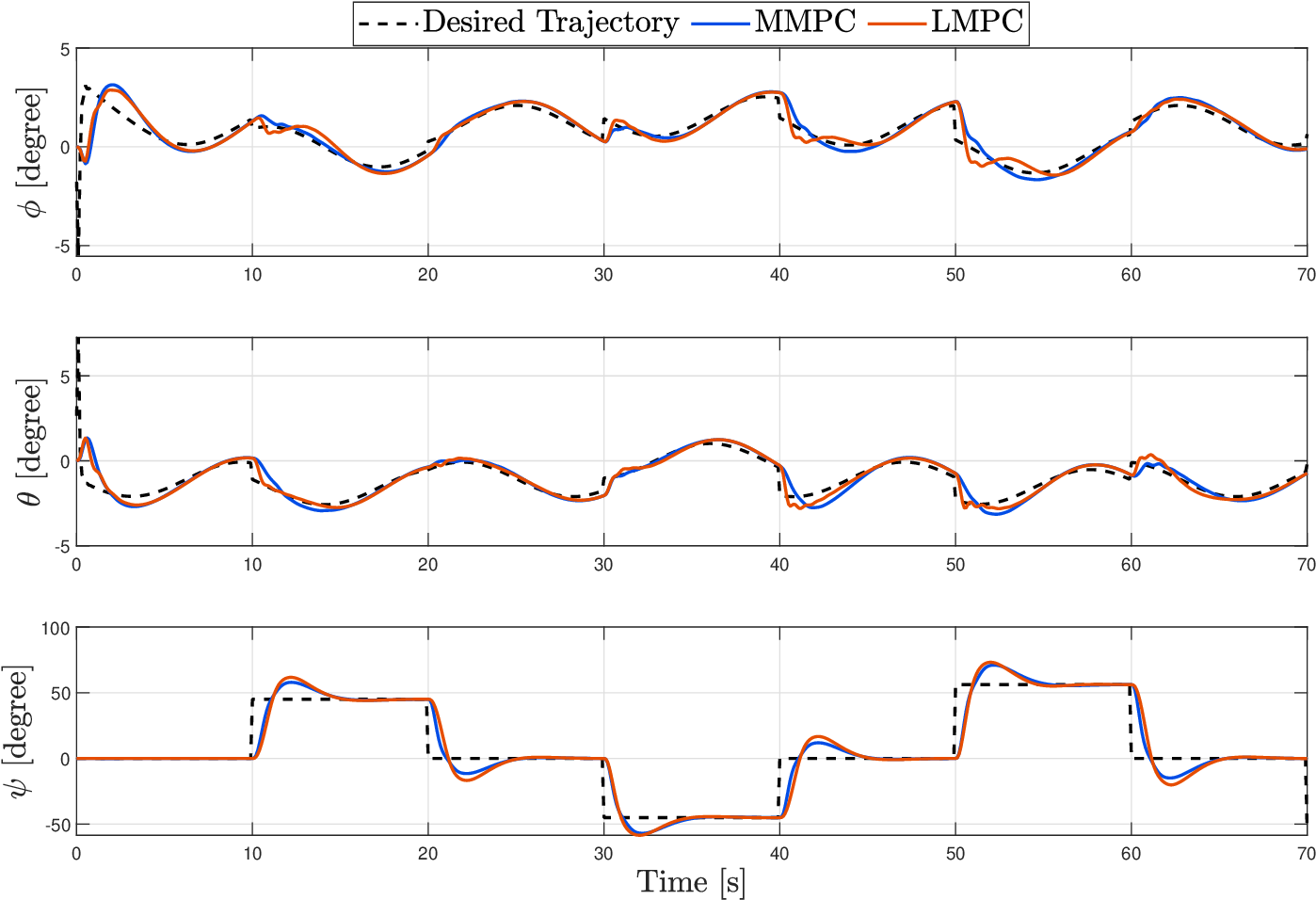}
    \caption{Quadrotor attitude trajectory in position control experiments}
    \label{fig:euler}
\end{figure}

\begin{figure}[t]
    \centering
    \includegraphics[width = 1\linewidth]{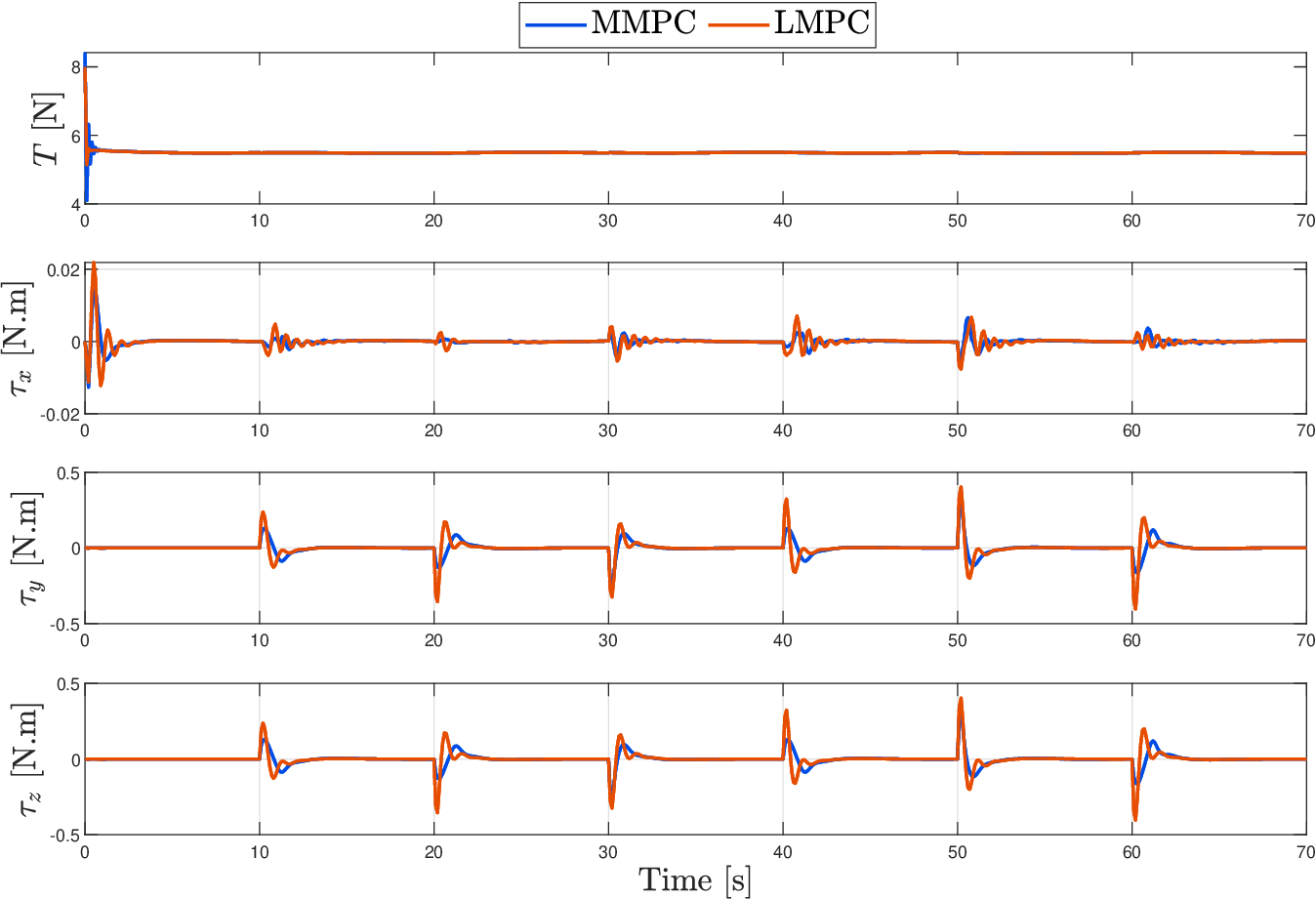}
    \caption{Quadrotor control signals in position control experiments}
    \label{fig:force}
\end{figure}

We compare the results for MMPC and LMPC, both working with identical SMC position controllers.
Figures \ref{fig:3D} -- \ref{fig:force} present the results.
The vehicle is capable of closely following the prescribed trajectory with both controllers; however, MMPC achieves more accurate position tracking than LMPC.

The reference roll and pitch angle generated by the position controller are shown in Fig. \ref{fig:euler} alongside the prescribed desired heading.
As expected, MMPC tracks the reference signals with higher accuracy compared to LMPC, and this, in turn, leads to improved position tracking of the vehicle.
In attitude control comparisons (Tab. \ref{tab:trajectory_tracking_error2}), it was revealed that MMPC uses smaller control effort compared to LMPC.
The same is evident in Fig. \ref{fig:force}, where the control signals of the controller with MMPC are significantly smaller than those with LMPC.

Overall, these results underscore the interconnected benefits of attitude and position control on the overall maneuverability of a quadrotor, indicating that improvements gained by MMPC in the attitude control loop positively affect the position control, leading to better performance and efficiency.



\begin{figure}[t]
    \centering
    \includegraphics[width = 1\linewidth]{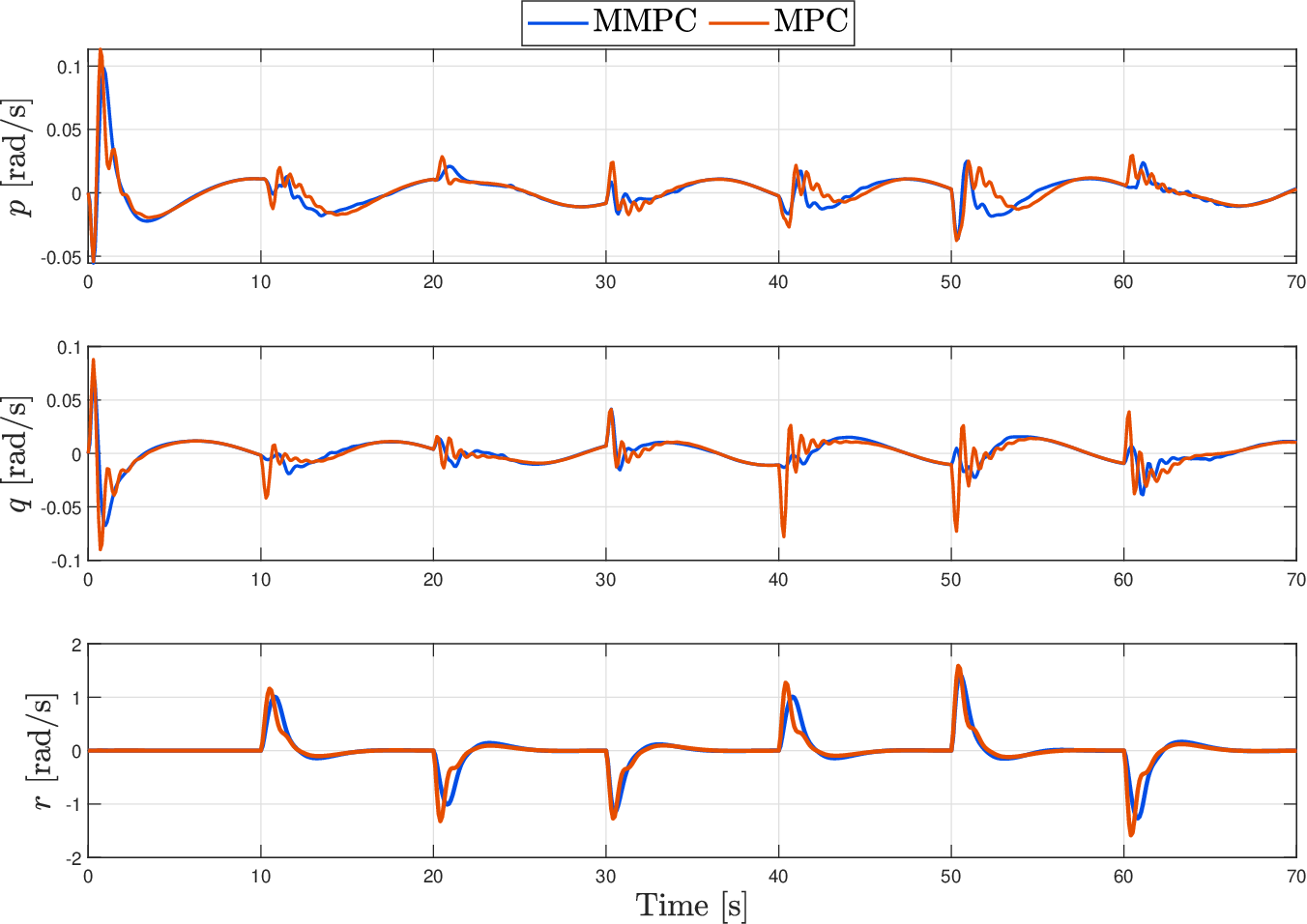}
    \caption{Quadrotor angular velocity in position control experiments}
    \label{fig:angular}
\end{figure}

\section{CONCLUSION}
This paper developed an MMPC for quadrotor attitude control.
The use of gap metric resulted in a significant reduction of models needed in the model bank, and the soft switching laws ensured system stability and performance.
Our results indicate that MMPC offers performance on par with NMPC but with a running time similar to LMPC.
This makes MMPC a compelling method for attitude control to guarantee a high refresh rate for resource-constrained quadrotors while maintaining accurate attitude tracking, ultimately, contributing to the enhanced position tracking and overall maneuverability of the vehicle.

Directions for future work include integration of MMPC with fast optimization solvers, hardware implementations, and integration with agile position controllers to unlock new potentials in agile flight control.
\bibliographystyle{IEEEtran}
\bibliography{reference}

\end{document}